\documentclass[10pt,journal,compsoc]{IEEEtran}

\IEEEoverridecommandlockouts

\usepackage{cite}
\usepackage{amsmath,amssymb,amsfonts}
\usepackage{algorithmic}
\usepackage{graphicx}
\usepackage{graphics}
\usepackage{textcomp}
\usepackage[export]{adjustbox}
\usepackage{multicol}
\usepackage{tabularx}
\usepackage{color}

\usepackage{hhline}
\usepackage{dblfloatfix}

\newcommand{\subparagraph}{}

\usepackage{titlesec}
\titleformat{\paragraph}
{\normalfont\normalsize\it}{\theparagraph}{0.5em}{}
\titlespacing*{\paragraph}
{0pt}{3.25ex plus 1ex minus .2ex}{1.5ex plus .2ex}

\usepackage[justification=centering]{caption}

\usepackage{enumitem}
\usepackage{comment}

\usepackage{hyperref}
\hypersetup{
    colorlinks=true,
    linkcolor=blue,
    filecolor=magenta,      
    urlcolor=red,
}
\def\BibTeX{{\rm B\kern-.05em{\sc i\kern-.025em b}\kern-.08em
    T\kern-.1667em\lower.7ex\hbox{E}\kern-.125emX}}

\begin{document}

\title{ Exploiting multi-CNN features in CNN-RNN based 
Dimensional Emotion Recognition on the OMG in-the-wild Dataset\\
}

\author{Dimitrios~Kollias and~Stefanos~Zafeiriou
\IEEEcompsocitemizethanks{\IEEEcompsocthanksitem D. Kollias and S.Zafeiriou are with the Department of Computing, Imperial College London, United Kingdom\protect\\
E-mail: dimitrios.kollias15@imperial.ac.uk \protect\\
%\IEEEcompsocthanksitem S. Zafeiriou is with the Department of Computing, Imperial College London, United Kingdom and with the Center for Machine Vision and Signal Analysis, University of Oulu, Oulu, Finland\protect\\
E-mail: s.zafeiriou@imperial.ac.uk}
%\thanks{Manuscript received February 15, 2019}
}

\markboth{IEEE Trans. on Affective Computing: Sp. I. on Automated Perception of Human Affect from Longitudinal Behavioral Data}%
{D. Kollias \MakeLowercase{\textit{et al.}}}

\IEEEtitleabstractindextext{%
\begin{abstract}

This paper presents a novel CNN-RNN based approach, which exploits multiple CNN features for dimensional emotion recognition in-the-wild, utilizing the One-Minute Gradual-Emotion (OMG-Emotion) dataset. Our approach includes first pre-training with the relevant and large in size, Aff-Wild and Aff-Wild2 emotion databases. Low-, mid- and high-level features are extracted from the trained CNN component and are exploited by RNN subnets in a multi-task framework. Their outputs constitute an intermediate level prediction; final estimates are obtained as the mean or median values of these  predictions. Fusion of the networks is also examined for boosting the obtained performance, at Decision-, or at Model-level; in the latter case a RNN was used for the fusion. Our approach, although using only the visual modality, outperformed state-of-the-art methods that utilized audio and visual modalities. Some of our developments have been submitted to the OMG-Emotion Challenge, ranking second among the technologies which used only visual information for valence estimation; ranking third overall. Through extensive experimentation, we further show that arousal estimation is greatly improved when low-level features are combined with high-level ones.
\end{abstract}

\begin{IEEEkeywords}
Deep convolutional and recurrent neural architectures; CNN plus Multi RNN; low-, mid-, high-level features; multi-CNN feature extraction and aggregation; multi-task learning; facial image analysis; valence; arousal; emotion recognition in-the-wild; AffWildNet; AffWild and AffWild2 emotion databases; OMG-Emotion database and Challenge. 
\end{IEEEkeywords}}

\maketitle

\ifCLASSOPTIONcompsoc
\IEEEraisesectionheading{\section{Introduction}\label{sec:introduction}}
\else
\section{Introduction}
\label{sec:introduction}
\fi

%%%%% sta au, va, expr kane cite edw
%%%%% stin last protasi me related research kane cite ta papers sou -> GENIKA KANE CITE TA PAPERS SOU

\IEEEPARstart{A}{utomatic}
 analysis of facial behaviour is the cornerstone of many application areas, including Human-Computer and Robot Interaction, Pervasive Computing, Ambient Intelligence and Virtual Reality. The research area of facial behaviour analysis includes the problems of: i) the recognition of the so-called six universal expressions (i.e., Anger, Disgust, Fear, Happy, Sad, Surprise), plus Neutral, influenced by the seminal work of Ekman\cite{ekman2003darwin}, ii) the recognition of spontaneous expressions including mental states (pain intensity \cite{kaltwang2012continuous} and compound expressions\cite{du2014compound}), iii) the detection of the facial Action Units (AU) and estimation of their intensity, according to the Facial Action Coding System \cite{ekman2002facial} which provides a standardised taxonomy of facial muscles' movements, iv) the detection of micro-expressions, and v) the estimation of facial affect in a continuous dimensional space (e.g., valence and arousal). Related research can assist in flagging complex behavioral patterns such as deception, depression, autism, spectrum disorders and schizophrenia \cite{tagaris1,tagaris2,kollias13,acharya2018automated,nasser2019artificial,kim2016deep}.

\begin{figure}[h]
\centering
\adjincludegraphics[height=5cm,width=5cm]{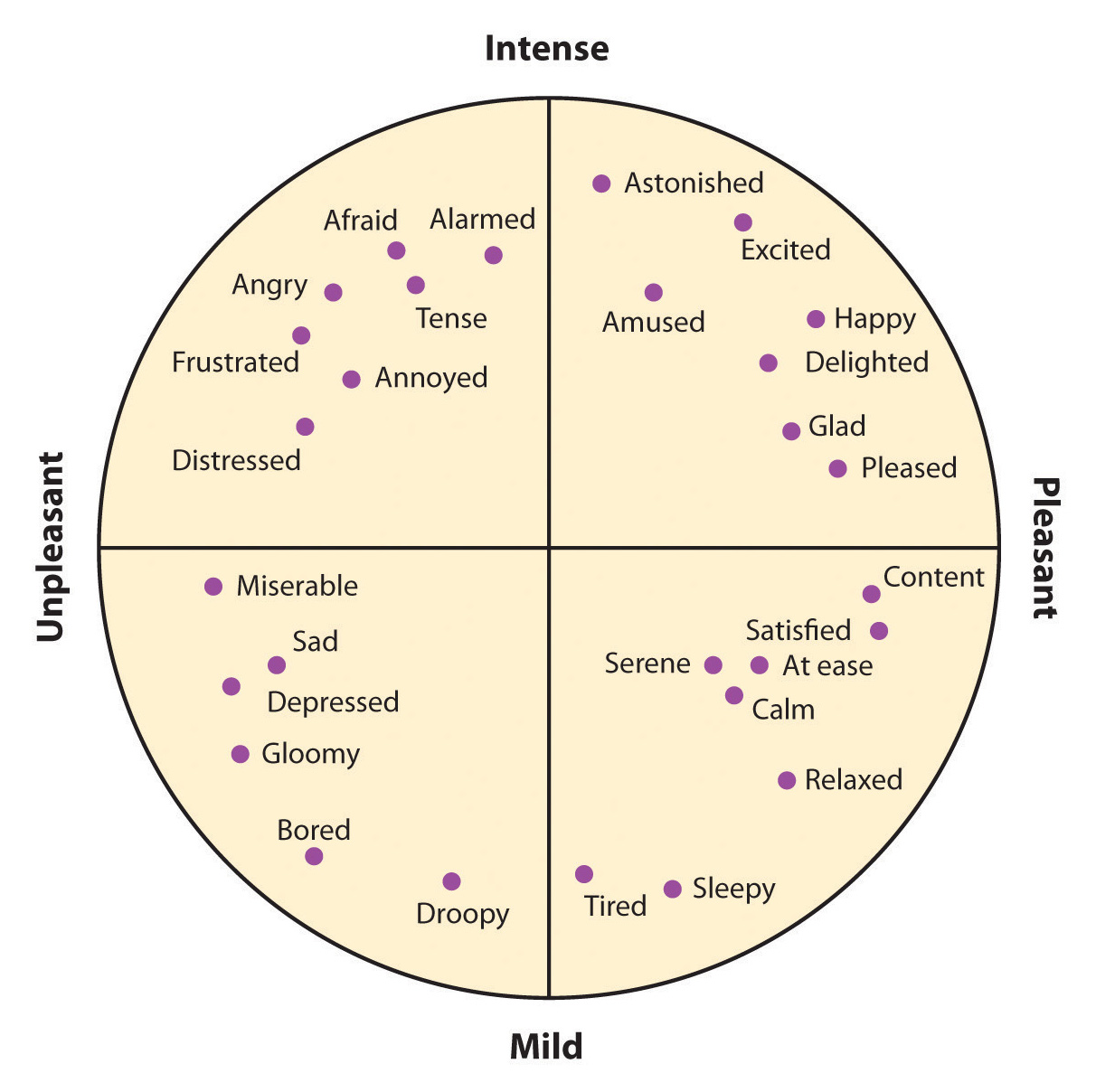}
\caption{The 2-D Valence-Arousal Space, as seen in \cite{little2012introduction}}
\label{2d-va-space}
\end{figure}

The main focus of this paper is on dimensional emotion models, which are appropriate to represent not only extreme, but also subtle emotions appearing in everyday human-computer interactions. According to the dimensional approach \cite{russell1978evidence} \cite{whissel1989dictionary}, affective behavior is described by a number of latent continuous dimensions. The most commonly used dimensions include valence (indicating how positive or negative an emotional state is) and arousal (measuring the power of emotion activation). Valence and arousal relate readily to specific functions of regions of the brain \cite{tom1990psychological,iordan2017brain,mickley2009effects}; the parietal region of the right hemisphere appears to play a special role in the mediation of arousal, whereas the frontal regions appear to play a special role in emotional valence. A third dimension, tension, is also introduced but often excluded due to difficulties in consistently identifying what the dimension describes: tension, control, or potency (dominance). 
Fig. \ref{2d-va-space} shows the 2-D Valence-Arousal Space, introduced in  \cite{plutchik1980emotion}. Estimation of valence and arousal continuous values related to affect constitutes the problem examined in the following.

In order to facilitate research on the above problems, many databases have been generated and annotated, most of which are in well-controlled conditions. %It has not been until recently that the focus has turned into ”in-the-wild” data. 
In the beginning, data and annotations
were scarce, hence research relied on extracting highly engineered handcrafted features and designing ad-hoc learning strategies \cite{meng2011naturalistic,glodek2011multiple,ramirez2011modeling,nicolle2012robust,meng2013depression}. Naturally, as the amount of data and annotations grew, research  has started to capitalise on data-intensive technologies, such as deep learning \cite{khorrami2016deep,chen2017multimodal,weichi,liu2018multi,kollias2015interweaving,kollias2020deep}.

It is now widely accepted, in both the computer vision and machine learning communities, that progress in a particular application domain is significantly catalysed when a large number of datasets are collected in unconstrained conditions (also referred as "in-the-wild" data). Hence, facial analysis could not only focus on spontaneous behaviors, but also on behaviors captured in unconstrained conditions. In-the-wild dimensional databases have been generated, %and annotated in terms of: i) basic expressions, such as the AFEW (audiovisual database) \cite{dhall2017individual}, the AffectNet (with static images) \cite{mollahosseini2017affectnet}, the RAF-DB (with static images) \cite{li2017reliable}, ii) action units, such as the EmotioNet (with static images) \cite{emotionet2016} and (iii) valence-arousal, 
such as the audiovisual OMG-Emotion Dataset \cite{barros2018omg}, Aff-Wild \cite{zafeiriou} \cite{kollias2018deep}, Aff-Wild2 \cite{kolliasexpression}\cite{kollias2018aff2}\cite{kollias2018multi}\cite{kollias2020analysing} and SEWA \cite{ringeval2017avec} ones, as well as AffectNet \cite{mollahosseini2017affectnet} which includes only static images. %and the AFEW-VA (audiovisual database)  \cite{kossaifi2017afew}.  

Regarding the pipeline of facial behavior analysis, the standard paradigm has been to: i) detect and/or track the face in an image sequence, ii) detect and/or track facial landmarks, iii) extract handcrafted features\footnote{Examples of handcrafted features include Histogram of Oriented Gradients (HoGs), Scale Invariant Feature Transform (SIFT), Local Binary Patterns (LBPs) and features from multiscale and multiorientation Gabor filterbanks}, either around the landmarks, or on the face region as a whole, and iv) use the features and the landmarks for classification/regression using affective labels. Recently this paradigm has shifted from utilizing handcrafted features to utilizing features learned by deep Convolutional Neural Networks (CNNs) and/or Recurrent Neural Networks (RNNs). This shift was motivated by the striking performance achieved when utilizing deep neural networks (DNNs) in a variety of emotion recognition tasks \cite{ding2017facenet2expnet,han2016incremental,ng2015deep,zhao2016peak,kollias6}.

In this paper, we address the issue of estimating  valence and arousal utilizing the One-Minute-Gradual  Emotion  Dataset  (OMG-Emotion Dataset), based on visual information only. We present novel deep neural architectures that provide best performance in valence and arousal estimation, as well as the submissions we made to the OMG-Emotion Challenge, which were ranked very high, especially for valence estimation.

The first main contribution of this paper is the development of CNN plus multi-RNN architectures for valence and arousal estimation in a multi-task optimization formulation. In this formulation, low-, mid- and high- level features are extracted from different layers of the CNN part and passed as input to the RNN part. The intuition for this is that these features include rich information which can be advantageous for the studied task.
%they contain semantic information as relevant for the}

These architectures are of two different types; in the first, the features extracted from, say, $K$ CNN layers are concatenated and passed as input to a single RNN, whereas in the other, they are passed to $K$ RNNs.
%each extracted feature is passed as input to a different RNN each time, so if $K$ features of dimensionality $R_i$ ($i \in \{1,..,K\}$) are extracted from $K$ layers of the CNN, they are passed as input to $K$ RNNs. 
In the experimental section, it is shown that the latter type outperformed all other developed architectures and even state-of-the-art networks that used not only the visual, but also the audio modality.
Our work deviates from others, such as \cite{liu2018multi,chen2017multimodal,deng2018multimodal},
that either: i) use standard CNN-RNN networks in which the output of the CNN is passed to the RNN, or ii) apply ensemble methodologies, using features extracted from many CNN networks (but not using features from multiple layers of the same network) and fusing them.

%\textcolor{red}{The difference between these types of architectures from others \cite{fan2018video,chen2017multimodal,deng2018multimodal} which extract features from intermediate levels is that the latter concatenate the extracted features and pass them through a series of convolutional and fully connected layers. Related works either do not use one RNN or they concatenate the features instead of passing each one through a series of new layers/through RNNs. + auto sto emotiw pou den kanei concatenate ta features alla apo kathe layer bgazei ena output k exei polles loss functions / isws periegrapse better k pio analutika ena ena ta papers + ta papers kanoun ensemble k pernoun features from different nets k den dokimasan na exoun ena net k na kanoun extract features apo auto}

%from each of the $N$ layers are passed as input to $N$ RNNs or differently the features extracted from each layer of the CNN are passed as input to a different RNN, meaning that if $N$ features of dimensionality $R_i, i = {1,..,n}$ are extracted from $N$ layers of the CNN, they are passed as input to $N$ RNNs. 

Both facial images and landmarks (after applying a Procrustes Analysis) are provided as inputs to these architectures. Additionally, ensemble formulations are proposed, using different levels of fusion (Model- or Decision-level) on the proposed architectures; these formulations are shown to further boost the obtained performance. In model-level fusion, our proposal is to perform fusion through a RNN instead of the typical fully connected layer.

Another contribution of this work is the approach to fit the developed architectures to the OMG-Emotion dataset characteristics and in particular to the dataset's annotation at utterance level. To deal with this, we split each utterance into sequences, which were individually processed by the above architectures. The mean or median of the predicted valence-arousal values were computed per sequence. Then, the means/medians were averaged at utterance level to provide the final valence and arousal estimates. This procedure deviates from related works that uniformly (or randomly) sample a constant number of frames from each utterance, assign to each of them the annotation value of the utterance and compute the prediction per frame \cite{deng2018multimodal}.

%split each utterance into sequences, randomly select one frame from each sequence (discarding the rest of the frames) and then pass these randomly selected frames through a CNN plus RNN with either temporal pooling \cite{peng2018deep} or attention mechanism \cite{zheng2018multimodal}

An additional contribution of this work is the pre-training of the proposed architectures on the large-scale emotionally rich Aff-Wild database and on its larger extension, the Aff-Wild2. Other works \cite{peng2018deep,zheng2018multimodal,triantafyllopoulos2018audeering} used networks that were not pre-trained on same task (valence-arousal estimation) but on other tasks (face recognition, object detection). The pre-training on these specific databases provided our developed architectures with the ability to effectively capture the dynamics of the OMG-Emotion in-the-wild dataset and thus provided a better performance.

The main findings of our approach have been: i) low-level features when combined with high-level ones in our CNN plus multi-RNN architectures, helped in boosting the networks' performance in arousal estimation; ii) CNN plus multi-RNN architectures outperformed standard CNN plus RNN ones showing that features extracted from previous layers contain useful and rich information for valence-arousal prediction; iii) better results were obtained when the features extracted from previous layers were processed by independent RNNs instead of being concatenated and fed to a single RNN; iv) better results were obtained when using a RNN instead of a fully connected layer for model-level fusion; v) when using the visual modality, network performance for valence estimation is much higher than the corresponding for arousal estimation.

The rest of this paper is organized as follows.
Section \ref{related-work} reviews related work and existing  state-of-the-art methods for facial expression recognition with emphasis on the dimensional model of affect. Section \ref{dbs} gives a brief description of the databases used in our experiments, i.e., OMG-Emotion Dataset, Aff-Wild and Aff-Wild2 databases. Section \ref{pre-processing} presents the pre-processing steps which were essential to obtain a common input representation for analysis.
Section \ref{methods} presents the developed methods, i.e., the created novel deep neural architectures, including ensembles and fusion of networks, for valence-arousal estimation. Section \ref{training} describes specific implementation details that we followed to achieve the best results. Section \ref{experiments} provides an evaluation of our approach by analysing the obtained results and presenting comparisons with other methods, in terms of achieved performance. Finally, Section \ref{conclusion} presents the conclusions.

\section{Related Work}\label{related-work}

One of the first deep learning architectures for valence and arousal estimation was proposed in \cite{khorrami2016deep}. In this work, both frame-based CNN and CNN plus RNN architectures were proposed and compared. The CNN consisted of 3 convolutional layers; the first two layers were followed by max pooling layers and the third by a quadrant pooling layer. A fully connected layer was then used, followed by the output layer. The CNN plus RNN architectures consisted of the previously described CNN network (keeping its weights fixed) without the top regression layer, followed by a single RNN layer that gave the final estimates. This methodology achieved very high valence and arousal correlations in a part of the RECOLA database \cite{ringeval2013introducing}. %RECOLA was created in controlled conditions and has been used in former AVEC Challenges \cite{valstar2016avec}.

The authors in \cite{chen2017multimodal} explored and
fused different hand-crafted and deep learning features from all available modalities  (acoustic, visual, and textual). They also considered the interlocutor influence (a person's influence on the interacting partner's behaviors) for the acoustic features. 

In more detail, the authors extracted: i) from the acoustic modality, hand-crafted features, such as MFCCs, loundness, F0, jitter, shimmer and features learned from the SoundNet \cite{aytar2016soundnet}, ii) from the visual modality, features learned from VGG-FACE \cite{parkhi2015deep} and DenseNet \cite{huang2017densely} that had been pre-trained on the FER+ \cite{BarsoumICMI2016} dataset (annotated in terms of the basic expressions), and iii) from the textual modality, word vectors that were used as features. All those features were fused and passed as input to a LSTM network that produced the estimates for valence, arousal and likability. This approach was the winning of AVEC 2017 Challenge that utilized the SEWA database. %, which records audiovisual spontaneous human to human interactions in the wild.   

The authors of \cite{weichi} presented the FATAUVA-Net method, which is a deep learning framework in which a core layer, an attribute layer, an AU layer and a valence-arousal layer were trained sequentially. The core layer was a series of convolutional layers, followed by the attribute layer which extracted facial area's features (face, eye, eyebrow, mouth). These layers were used in supervised learning of AUs. Finally, AUs were employed as mid-level representations to estimate the intensity of valence and arousal. This methodology produced the highest results of the First Affect-in-the-wild Challenge\cite{kollias2018deep} which was the first  challenge on the estimation of valence and arousal in-the-wild, using the Aff-Wild database for recognition of affect.

Best results in the Aff-Wild database have been obtained by the authors of \cite{zafeiriou1}\cite{kollias2018deep}. In these works, the authors performed a large number of experiments, training CNN and CNN-RNN networks on the Aff-Wild for emotion recognition. The best performing network, AffWildNet, consists of the convolutional and pooling parts of the ResNet-50 network \cite{he2016deep}, followed by a fully connected layer, a 2-layer GRU \cite{chung2014empirical} and the output layer that provided the final valence-arousal estimates. This network was further fine-tuned on the RECOLA and AFEW-VA databases, producing state-of-the-art performance.
%, both during the Challenge and in the following, producing best performances \cite{zafeiriou1}\cite{kollias2018deep}. Moreover, as shown in \cite{kollias2018deep}, best performances were obtained over RECOLA and AFEW-VA as well, using the AffWildNet network consisting of the convolutional and pooling parts of the ResNet-50 network \cite{he2016deep}, followed by a fully connected layer, a2-layer GRU \cite{chung2014empirical} and the output layer that provided the final valence-arousal estimates.}

Table \ref{rel-work} provides a summary of the performance of the above-described methods on the respective databases.

\begin{table}[h]
\caption{State-of-the-art algorithms for valence-arousal estimation, their performances and utilized databases}
\label{rel-work}
\centering
\scalebox{0.75}{
\begin{tabular}{ |c|c|c|c| }
\hline
 Work & Databases Used & Methods & Results  \\
   \hhline{=:=:=:=}
 \cite{khorrami2016deep} & \begin{tabular}{@{}c@{}} part of \\ RECOLA \\ as used in the \\ AVEC Challenge \end{tabular} & \begin{tabular}{@{}c@{}} CNN-RNN visual only: \\ (conv + max-pool) x2 \\ + conv + quadrant-pool \\ + RNN \end{tabular} & \begin{tabular}{@{}c@{}} Valence:  \\ RMSE = 0.107 \\ PCC = 0.554 \\ CCC = 0.507 \end{tabular}  \\
 \hline
\cite{chen2017multimodal} & SEWA & \begin{tabular}{@{}c@{}} (1) audio: \\ handcrafted + SoundNet features \\ (2) visual: \\  VGG-FACE + DenseNet features \\ (3) text: \\ word vectors - features \\ fusion of (1), (2), (3) + LSTM \end{tabular}  & \begin{tabular}{@{}c@{}} Valence - Arousal:  \\ RMSE = 0.081 - 0.086 \\ PCC = 0.758 - 0.702 \\ CCC = 0.756 - 0.672 %\\  Arousal:  \\ RMSE = 0.086  \\ CCC = 0.672 
\end{tabular} \\ 
 \hline
\cite{weichi} & Aff-Wild & \begin{tabular}{@{}c@{}} 1) core layer: series of conv. layers \\ 2) attribute layer: facial features  \\ 3) AU layer \\  4) Valence \& Arousal layer \end{tabular}  & \begin{tabular}{@{}c@{}} Valence - Arousal  \\ MSE = 0.123 - 0.095 \\ CCC = 0.396 - 0.282 %\\ Arousal: \\ MSE = 0.095 \\ CCC = 0.282 
\end{tabular} \\ 
 \hline
\begin{tabular}{@{}c@{}}\cite{zafeiriou1} \\ \cite{kollias2018deep} \end{tabular} & \begin{tabular}{@{}c@{}} Aff-Wild; \\ whole RECOLA; \\AFEW-VA; \end{tabular} & \begin{tabular}{@{}c@{}} AffWildNet: \\ ResNet-50 + FC + GRU  \end{tabular}   & \begin{tabular}{@{}c@{}}  CCC: \\ Valence - Arousal  \\ Aff-Wild : 0.570 - 0.430 \\ RECOLA :  0.526 - 0.273 \\ AFEW-VA: 0.515 - 0.556  \end{tabular}   \\
 \hline
\end{tabular}
}
\end{table}

%It is this developed network that we adopted as a pre-trained, starting point for deriving the DNN architecture used to tackle the OMG Challenge and achieve the second best position in valence estimation, when using only visual inputs, or the third best results, when all multi-modal inputs were used. In this paper, we also show that using a recent greatly expanded version of the Aff-Wild database, named Aff-Wild2, and the proposed architecture, we are able to further improve this performance on the OMG dataset. 

\section{The utilized in-the-wild Dimensional Emotion Databases }\label{dbs}

\begin{figure*}[h]
\centering
\adjincludegraphics[height=1.5cm,width=1.5cm]{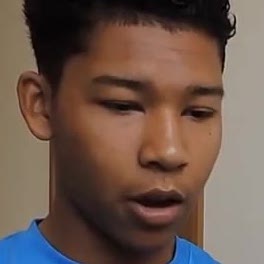}
\adjincludegraphics[height=1.5cm,width=1.5cm]{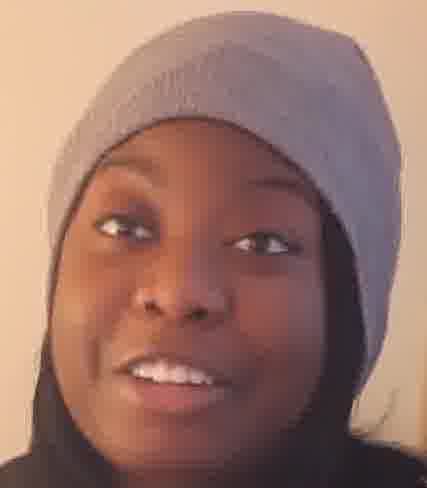}
\adjincludegraphics[height=1.5cm,width=1.5cm]{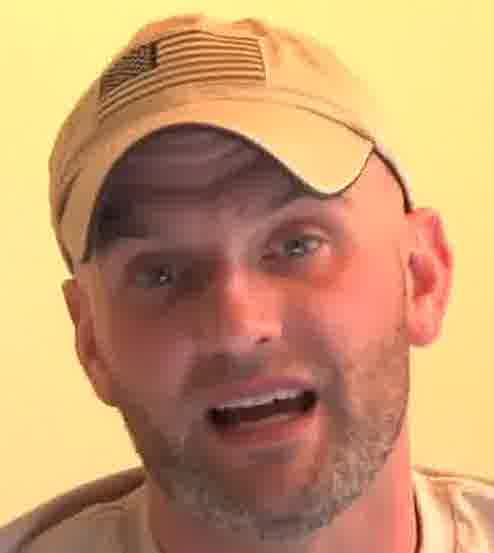}
\adjincludegraphics[height=1.5cm,width=1.5cm]{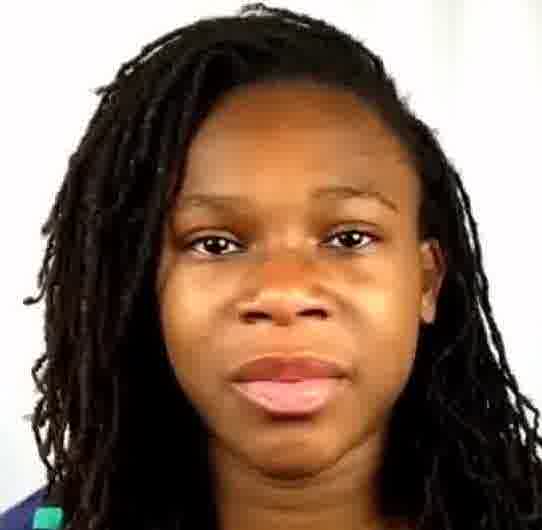}
\adjincludegraphics[height=1.5cm,width=1.5cm]{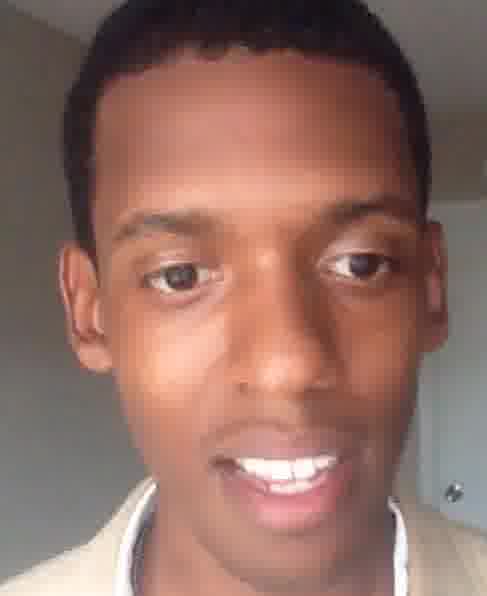}
\adjincludegraphics[height=1.5cm,width=1.5cm]{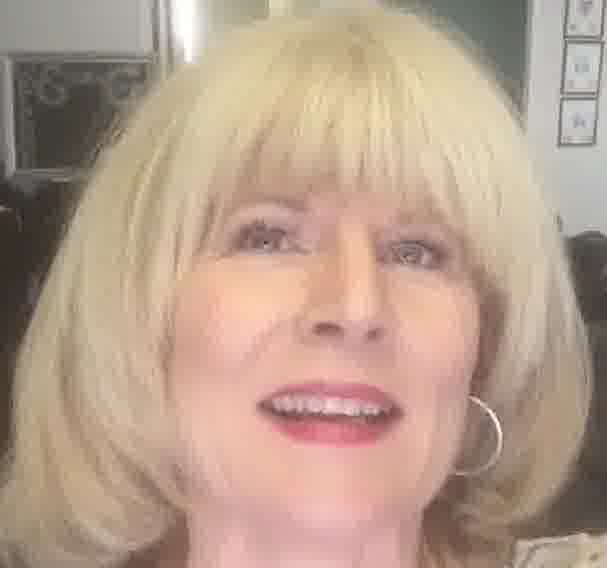}
\adjincludegraphics[height=1.5cm,width=1.5cm]{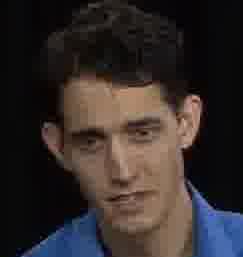}\\
\adjincludegraphics[height=1.5cm,width=1.5cm]{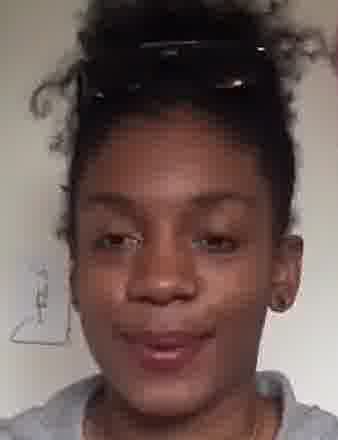} 
\adjincludegraphics[height=1.5cm,width=1.5cm]{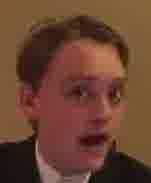}
\adjincludegraphics[height=1.5cm,width=1.5cm]{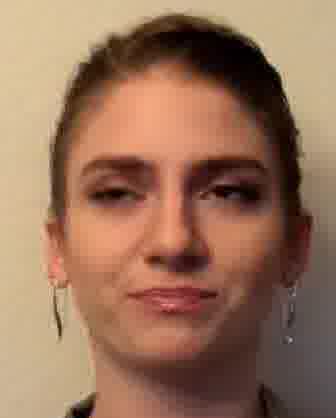}
\adjincludegraphics[height=1.5cm,width=1.5cm]{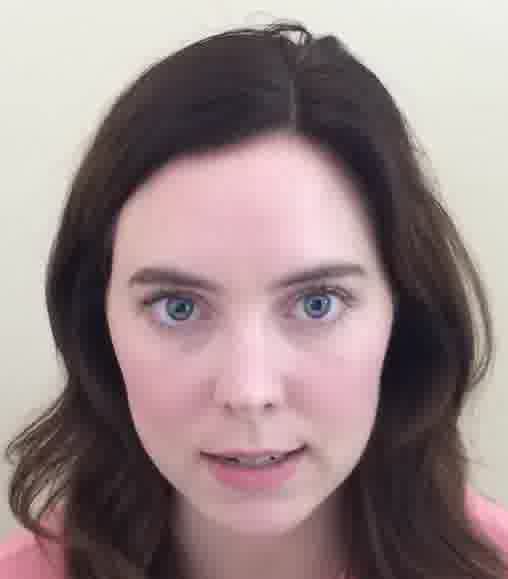}
\adjincludegraphics[height=1.5cm,width=1.5cm]{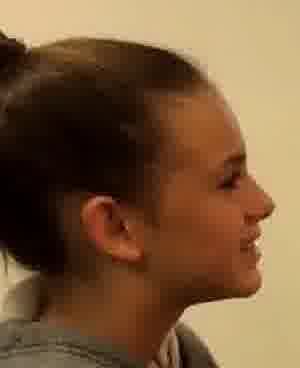}
\adjincludegraphics[height=1.5cm,width=1.5cm]{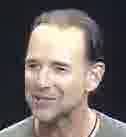}
\adjincludegraphics[height=1.5cm,width=1.5cm]{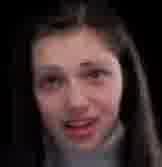}
\caption{Sample images from the OMG-Emotion dataset showing people displaying various in-the-wild emotions}
\label{omg-samples}
\end{figure*}

\begin{figure*}[!b]
\centering
\adjincludegraphics[height=3cm,width=4.5cm]{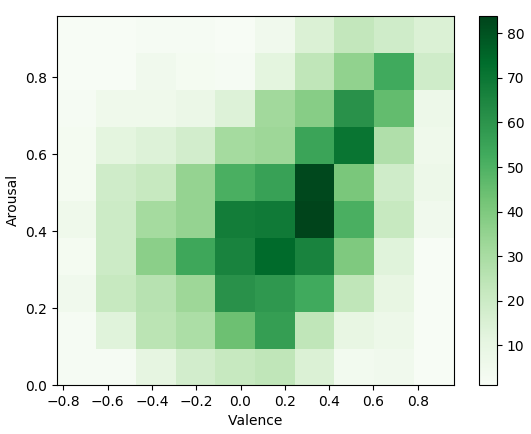}
\adjincludegraphics[height=3cm,width=4.5cm]{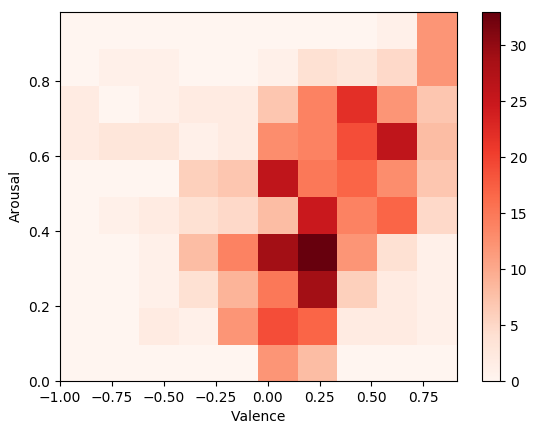}
\adjincludegraphics[height=3cm,width=4.5cm]{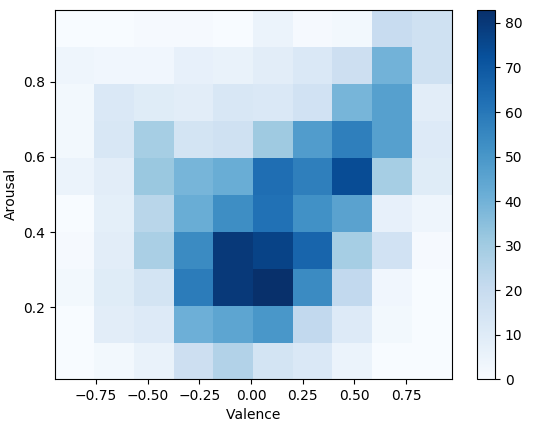}
\\ 
\adjincludegraphics[height=3cm,width=4.5cm]{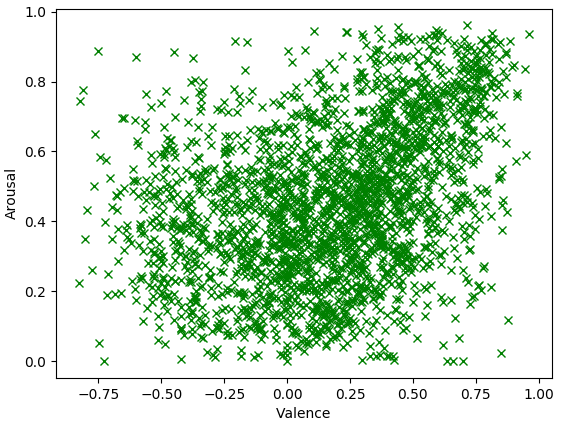}
\adjincludegraphics[height=3cm,width=4.5cm]{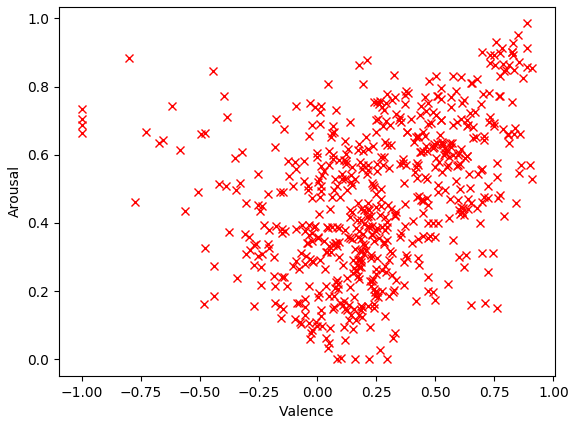}
\adjincludegraphics[height=3cm,width=4.5cm]{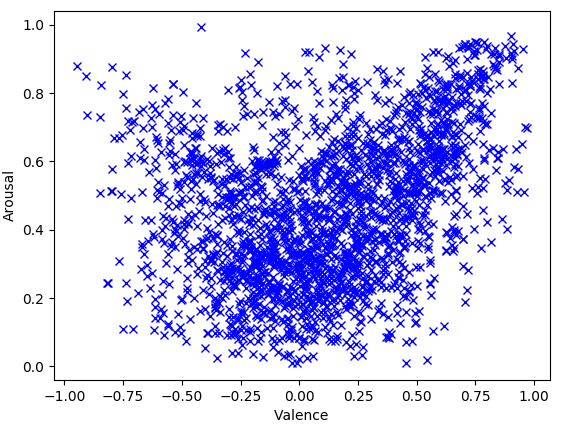}
\caption{Histograms (on the top row) and Distributions (on the bottom row) in the 2-D Valence-Arousal space of utterance-level annotations of the training (in green), validation (in red) and test (in blue) OMG-Emotion sets}
\label{omg-hist-data}
\end{figure*}

In this Section we provide a short description of the Aff-Wild and its extension Aff-Wild2 database, which have been used to pre-train the developed deep neural network architectures, as well as the OMG-Emotion database, analysis of which is the target of this paper.  

\subsection{Aff-Wild Database}

The Aff-Wild database \cite{kollias2018deep} \cite{zafeiriou} has been the first large scale captured in-the-wild database that has been annotated by 8 lay experts with regards to valence and arousal. It consists of 298 videos and displays reactions of 200 subjects, with a total video duration of more than 30 hours. The total number of frames in this database is 1,224,100. Regarding subjects' gender, 130 are male and 70 female. The Aff-Wild database served as benchmark for the Aff-Wild Challenge, organized in conjunction with CVPR 2017. The aim for this database was to collect spontaneous facial behaviors in arbitrary recording conditions. To this end, the videos were collected using Youtube. The main keyword that was used to retrieve the videos was ”reaction”. 

\subsection{Aff-Wild2 Database}

The Aff-Wild database has recently been augmented with new Youtube videos having a total length of $13$ hours and $5$ minutes, thus forming the Aff-Wild2 database \cite{kolliasexpression,kollias2018aff2,kollias2018multi}. This database has been the basis for the ABAW Competition \cite{kollias2020analysing}. The aim has been to extend the spontaneous facial behaviors in arbitrary recording conditions  met  in  Aff-Wild, whilst significantly increasing the number of different subjects in it. All the additional videos have been annotated by four experts and contain a wide range in subjects': age (from babies and young children, to elderly people); ethnicity (subjects are caucasian, hispanic or latino, asian, black, or african american); profession (e.g. actors, athletes, politicians, journalists); head pose; illumination conditions; occlusions; emotions. In  total, Aff-Wild2  consists  of  558  videos  with 2,786,201 frames. $11$ out of those videos display two subjects, all of which have been annotated. The total number of subjects is 458, with 279 of them being male and 179 female.

\subsection{OMG-Emotion Dataset}\label{omg-db}

The One-Minute Gradual-Emotional Behavior dataset
(OMG-Emotion dataset) \cite{barros2018omg} contains in-the-wild videos from Youtube where emotion expressions emerge and develop over time based on monologued scenarios. Figure \ref{omg-samples} shows some frames from this dataset, with various people displaying different emotions under many occasions/circumstances.
This dataset is annotated in terms of valence and arousal and also contains a large number of different identities.

The OMG-Emotion dataset served as a benchmark for the One-Minute Gradual-Emotion Recognition (OMG-Emotion) Challenge \footnote{\url{https://www2.informatik.uni-hamburg.de/wtm/omgchallenges/omg_emotion2018_session.html}},  held jointly with the Special Session on Neural Models for Behavior Recognition at the WCCI/IJCNN 2018. In particular, the dataset is split into training, validation and test sets in a subject independent manner, meaning that each subject appears strictly in only one of these sets. The training set consists of 231 videos composed of 2442 utterances, the validation set consists of 60 videos composed of 617 utterances and the test set consists of 204 videos composed of 2229 utterances. Each utterance has an average length of 8 seconds and each video has an average length of around 1 minute.

For annotating the collected data, the Amazon Mechanical Turk tool was used, resulting, on average, in five independent annotations per utterance. Each annotator was given the full contextual information of the video up to that point when annotating the dataset. That means that each annotator could take into consideration not only the visual and audio information but also the context of each video, i.e. what was spoken in the current and previous utterances through the context clips provided by the annotation tool. In this manner, each annotation is based on multimodal information.

Each utterance was given a specific valence and arousal value, based on the gold standard of the five annotations. Valence annotations range in $[-1,1]$, whereas arousal ones range in $[0,1]$. In Fig. \ref{omg-hist-data}, on top row are the 2-D histograms of the annotations in the OMG-Emotion training, validation and test sets, respectively, and in the bottom row are the corresponding datasets' annotations' distributions.

Additionally, this dataset contains categorical annotations for each utterance; transcripts of what was spoken in each of the videos are also provided.

\section{Pre-processing: Face Detection \& Alignment, Image Resizing \& Normalization}\label{pre-processing}

Data pre-processing consists of all processing steps that are required for starting the extraction of meaningful features from the data. The usual steps are face detection, face alignment, image resizing and image normalization.
%\subsection{Face Detection}
The first step is to extract face bounding boxes from all video frames. In order to do so, we used the Deformable Part Model (DPM) detector ffld2 \cite{mathias2014face}  that has proven to be highly efficient and accurate for face detection in-the-wild. %This weakly-supervised DPM detector is trained using only the bounding boxes of the positive examples and a set of negative examples. 

For face alignment, we extracted facial landmarks and implemented the Generalized Procrustes Analysis \cite{gower1975generalized}. %Facial landmarks are defined as distinctive face locations, such as the corners of the eyes, centre of the bottom lip, or the tip of the nose. If they are aggregated in sufficient numbers, they can effectively describe the face shape.
In our implementations, we first used the facial landmark detector inside the dlib library \cite{kazemi2014one} to locate 68 facial landmarks in all frames.
We used as reference and rigid points, 5 anchor points that corresponded to the location of the left eye, right eye, nose and mouth in a prototypical frontal face. For every frame, we used its 5 facial landmarks corresponding to the location of the same facial components; we performed Procrustes transformation, which eliminates in-plane rotation, isotropic scaling and translation, on the coordinates of these 5 landmarks and the coordinates of the 5 landmarks of the frontal face; we imposed this transformation to the whole new frame to perform the alignment.
All cropped and aligned images were then resized
to $96 \times 96 \times 3$ pixel resolution and their 
intensity values were normalized to the range $[-1, 1]$. Those images, along with the 68 facial landmarks, were then used as inputs for training our networks, as described in the following Section.

\section{The Developed Architectures}\label{methods}

This section presents the proposed framework for dimensional emotion recognition, by describing the CNN, the standard CNN plus RNN, the proposed CNN plus Multi RNN architectures and then an ensemble methodology for fusion. 

In all architectures presented in this Section, we compared a uni-task learning approach, independently for valence and arousal, to multi-task learning approach. The latter provided better performance in estimation of both affective dimensions. This result is in agreement with  \cite{nicolaou2013correlated} which claims that there exist inter-correlations between the valence and arousal emotion dimensions. %In other words, each emotion dimension can have higher correlation with the other dimension than with the audio or facial features.
This relation between emotion dimensions in isolation, (i.e., without including features), has been well-supported by related
research in psychology \cite{lane1999cognitive}   \cite{kuppens2017relation}. That is the reason why in the following, we focus on the multi-task case.

\subsection{CNN architectures}

\begin{figure*}[h]
\centering
\adjincludegraphics[height=3.8cm,width=15cm]{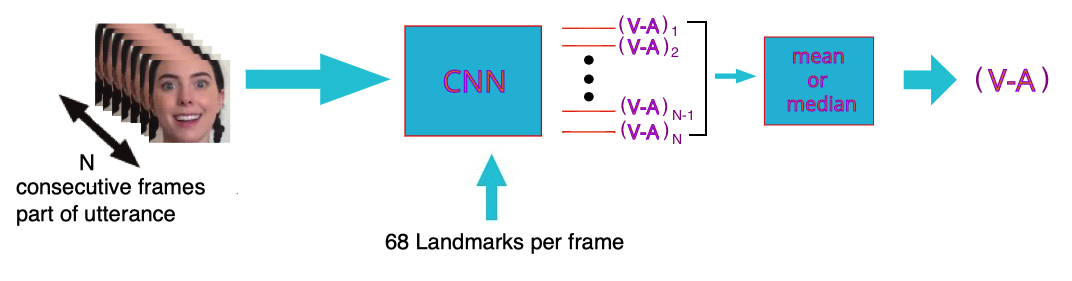}
\caption{The developed CNN structure. It gives only one valence-arousal (V-A) estimate per input sequence of consecutive frames. The CNN component can be any of the VGG-FACE, ResNet-50 and DenseNet-121 networks. The 68 landmarks are concatenated with the extracted features from the last pooling layer of the CNN component and are passed to the fully connected layer that precedes the output layer.}
\label{cnn}
\end{figure*}

We experimented with three state-of-the-art networks: VGG-Face, ResNet-50 and DenseNet-121. These networks were first pre-trained either on the Aff-Wild or the Aff-Wild2 database and then trained on the OMG-Emotion training set. %\textcolor{red}{The VGG-Face has proven to provide the best results, as reported in the experimental Section \ref{experiments}}.
To design the structure of these networks, we took into account the procedure used to annotate the OMG-Emotion dataset. According to this, each utterance was labeled with a single pair of valence and arousal values. We split each utterance into smaller parts-sequences, each consisting of the same number of consecutive frames. Then, we assigned to each of those parts-sequences of frames, the label of the corresponding utterance.

Training of the CNN networks was performed as shown in Fig.\ref{cnn}. In more detail, each CNN was provided with an input sequence and was trained to predict, for each frame in the sequence, the respective valence-arousal pair of values. The 68 facial landmarks (per each frame of the input sequence) were also provided as additional inputs to the CNN networks.
The final valence (arousal) prediction was computed as the mean, or median (both approaches were considered) of the per-frame valence (arousal) values in that sequence.

In Fig.\ref{cnn}, the CNN structure can be any of the VGG-FACE, ResNet-50 and DenseNet-121 ones.
In the VGG-FACE CNN case, the landmarks were concatenated with the outputs of the last pooling layer of the network and were given as input to the first fully connected layer, that consisted of 4096 units. In this way, both outputs and landmarks were mapped to the same feature
space, before performing the prediction.
%%%% isws prosthese tables gia tis 3 CNN architectures
In the ResNet-50 (and DenseNet-121) case, the landmarks were concatenated with the averaged pooled features of the ResNet-50 (DenseNet-121) network and were given as input to a fully connected layer consisting of 1500 units. This layer was followed by the output layer which provided the final estimates for valence-arousal pair.

\subsection{Standard CNN plus RNN architectures}\label{cnnplusrnn}

In order to consider the contextual information in the data and more specifically the temporal dependencies of facial expressions in each utterance, we designed standard CNN plus RNN architectures. In the following we present the different CNN-RNN architectures that we have developed and used in the experimental study. In these architectures, the output of the CNN's last pooling layer is being fed to a fully connected layer, whose output constitutes the input of the RNN layers.
These architectures were pre-trained on either the Aff-Wild, or the Aff-Wild2 databases. We then used two different strategies for training these architectures: i) keeping the CNN weights fixed and training the remaining architecture (i.e., the fully connected layers and the RNNs), or ii) training the whole architecture in an end-to-end manner (by jointly training the CNN and RNN parts). The latter approach provided the best results.

%\textcolor{red}{In the following we present the different CNN-RNN architectures that we have developed and used in the experimental study. We present these in 4 categories, namely CNN-RNN, CNN-1RNN, CNN-2RNN and CNN-3RNN networks. In the CNN-RNN category, the output of the CNN's last pooling layer is being fed to a fully connected layer, whose output constitutes the input of the RNN layers. In all other categories, we attempt to insert more information from the CNN part to the RNN part, providing it with more features, extracted from a lower (semantic) level.} 

%\textcolor{red}{Since CNN-1RNN provided the best arousal predictions and the CNN-3RNN the best valence predictions, they are presented in more detail.  }  

%\subsubsection{CNN-RNN networks}

\subsubsection{AffWildNet}

At first, we considered the AffWildNet \cite{kollias2018deep} as the best performing network on the Aff-Wild database and re-trained it on the OMG-Emotion database. As shown in Table \ref{affwildnet}, the AffWildNet is a CNN-RNN network consisting of the convolutional and pooling parts of ResNet-50 followed by a fully connected layer of 1500 units, followed by a 2-layer GRU, with each layer having 128 units. In this architecture, the landmarks are concatenated with the averaged pooled features of the ResNet-50 and being fed as input to the fully connected layer consisting of 1500 units. Similarly to the CNN case described in the previous Subsection, the CNN-RNN network receives an input sequence of frames, then predicts, for each frame, the valence-arousal values and finally computes the mean, or median, of these values, which is the final estimate. This architecture is the same as in Fig.\ref{cnn}, if one replaces the CNN network with the CNN-RNN (i.e., the AffWildNet).

\begin{table}[h]
\caption{The AffWildNet architecture}
\label{affwildnet}
\centering
\begin{tabular}{|c|c|c|}
\hline
block 1 & \begin{tabular}{@{}c@{}}  ResNet-50 conv \& pooling parts \end{tabular} & \\
\hline
block 2 &fully connected 1 & 1500 \\
&dropout &\\
\hline
block 3 & GRU layer 1 & 128\\
&dropout &\\
\hline
block 4 & GRU layer 2 & 128\\
\hline
block 5 &fully connected 2 &2\\
\hline
\end{tabular}
\end{table}

\subsubsection{DenseNet-RNN}

We also used a DenseNet-RNN structure that is quite similar to that of the AffWildNet described in the previous Subsection. The only difference is that it uses the DenseNet-121 network's convolutional and pooling layers.

\subsection{CNN plus Multi-RNN networks}\label{multi-rnn}

In general, features extracted from the low CNN layers contain rich, complete and time varying information, whilst high-level features are highly specific and characteristic of the specific problem studied.  Taking this into account, we have developed and used CNN plus Multi-RNN networks; these networks extract low-, mid-  and high- level features from different layers of the CNN and pass them through RNNs. These networks are split into two different types through different methodologies: the first, referred as CNN-1RNN, concatenates the extracted features from 3 CNN layers and passes them to a single RNN, whereas the other, referred as CNN-3RNN, processes them independently through 3 RNN subnets. %\textcolor{red}{Both types have the same complexity in terms of feature input in terms of big O/ $\mathcal{O}$ notation.}

It should be mentioned that we also tested other networks: CNN-2RNN (extracting features from 2 CNN layers and pass them independently to 2 RNNs); CNN-2RNN-1FC (similarly as before, with the outputs from the 2 RNNs being concatenated and passed to a fully connected layer; in this way they are both mapped to the same feature space, before performing the final prediction); CNN2-to-1RNN (extracting features from 2 CNN layers, concatenating them and passing them as input to a single RNN); CNN-3RNN-1FC (the outputs from the 3 RNNs being concatenated and passed to a fully connected layer, before performing the final prediction). These architectures provided performance that was around 4-5\% lower than the performance of the CNN-1RNN and CNN-3RNN networks, presented next in this Section.

%In the following we present the different CNN-RNN architectures that we have developed and used in the experimental study. We present these in 4 categories, namely CNN-RNN, CNN-1RNN, CNN-2RNN and CNN-3RNN networks.  In all other categories, we attempt to insert more information from the CNN part to the RNN part, providing it with more features, extracted from a lower (semantic) level.

%enrich the feature map of each layer, by combining the connections across the side-output layers.
%enhance the information from info from previous layers
%Features from the last convolutional layer and fully connected layers are highly specific, being either characteristic of a class or irrelevant for it. Features from the rest of convolutional layers convey more variate information, and can be characteristic of a class both by their presence or by their absence

\subsubsection{CNN-3RNN networks}\label{cnn_3rnn1_net}

The CNN-3RNN networks include the convolutional and pooling layers of VGG-FACE, followed by a fully connected layer of 4096 units. The 68 facial landmarks are concatenated with the features extracted from the last pooling layer of VGG-FACE and are passed to this fully connected layer. Then, low-, mid-  and high-level features are extracted and each one is processed by a 2-layer GRU network that predicts the valence and arousal values. Each GRU layer comprises 128 units. 
Similarly to the architectures described in Subsection \ref{multi-rnn}, the CNN-3RNN networks are provided with an input sequence of frames (and the corresponding landmarks of each frame), predicting, for each frame, the valence-arousal values; their mean, or median constitute the final estimates.

\begin{figure}[ht]
\centering
\adjincludegraphics[height=14.8cm,width=5cm]{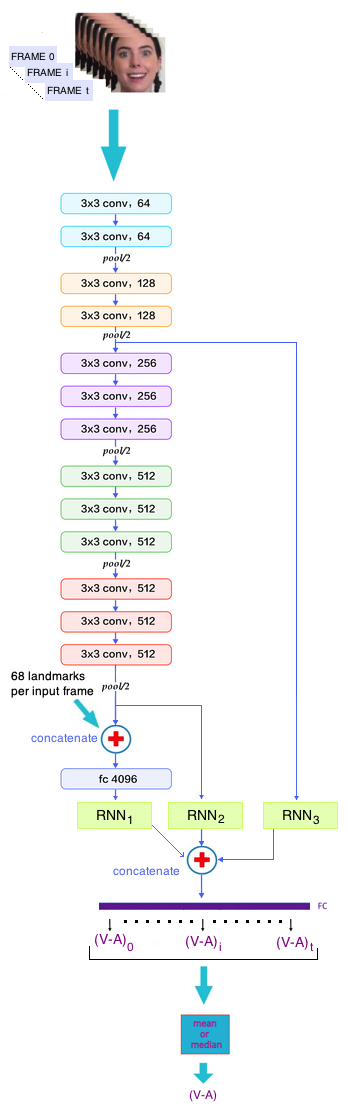}
\caption{The CNN-3RNN-2nd-pool\_last-pool\_fc. It provided a valence-arousal (V-A) estimate per input sequence of consecutive frames. The '68 landmarks' are concatenated with the features of the last 'pool' layer and passed as input to the 'fc' layer. This architecture provided the best results.}
\label{cnn_3rnn1}
\end{figure}

\begin{figure}[ht]
\centering
\adjincludegraphics[height=2cm,width=6cm]{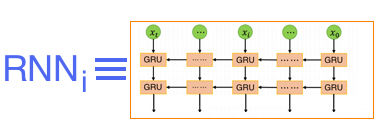}
\caption{The structure of each RNN network in the CNN-3RNN architecture displayed in Fig. \ref{cnn_3rnn1}.}
\label{cnn_3rnn2}
\end{figure}

Fig.\ref{cnn_3rnn1}, presents an example of CNN-3RNN networks, named CNN-3RNN-2nd-pool\_last-pool\_fc. In this network: i) the features extracted from  the fully connected layer are passed as input to a RNN network, denoted $RNN_1$ in Fig.\ref{cnn_3rnn1}; ii) the features extracted from the last pooling layer (before being concatenated with the landmarks) are passed as input to a second RNN network, denoted $RNN_2$ in Fig.\ref{cnn_3rnn1}; iii) the features extracted from the second pooling layer (following the fourth convolutional layer) are passed as input to another RNN network, denoted $RNN_3$ in Fig.\ref{cnn_3rnn1}. Fig.\ref{cnn_3rnn2} depicts the exact structure of the afore-mentioned $RNN_i$, $i \in\{1,2,3\}$, networks. All networks have the same structure; a 2-layer GRU network, with each layer having 128 units. Next, the outputs of the 3 RNNs are concatenated and passed to the output layer that performs the valence-arousal prediction.
As shown in the experimental Section \ref{experiments}, this network 
based on the features extracted from these specific layers provided the best results in these type of networks.

%The reason why we tested this was because that the output of the last pooling layer and that of the last convolutional layer were similar in the sense that the pooling layer subsamples

%\paragraph{CNN-3RNN-1FC network}
%In the CNN-3RNN-Variant 1 case described above, we also tested passing first the concatenated outputs of the 3 RNNs to a fully connected layer consisting of 128 units followed by the output layer.

\subsubsection{CNN-1RNN networks}\label{cnn_1rnn_net}

The CNN-1RNN types of networks consist of the convolutional and pooling layers of VGG-FACE, followed by a fully connected layer of 4096 units. The 68 facial landmarks are concatenated with the features extracted from the last pooling layer of VGG-FACE and are passed to this fully connected layer. Then, low-, mid-  and high-level features are extracted, concatenated and passed to a 2-layer GRU network that predicts the valence and arousal values. Each GRU layer comprises 128 units. 
Similarly to the other architectures described above, the CNN-1RNN networks are provided with an input sequence of frames (and the corresponding landmarks of each frame), predicting, for each frame, the valence-arousal values; their mean, or median, are the final estimates.

%Next, we describe these types of networks that extract 3 different CNN features from low- and high- levels and concatenate them before performing the final predictions. Different combinations of low- and high-level features were tested. The best combinations were if the features were extracted from convolutional layers  

%The features extracted from the fully connected layer are concatenated with the features extracted from the last pooling layer of VGG-FACE and are passed to a 2-layer GRU network that predicts the valence and arousal values. Each GRU layer comprises of 128 units. 

\begin{figure}[ht]
\centering
\adjincludegraphics[height=14.7cm,width=3.5cm]{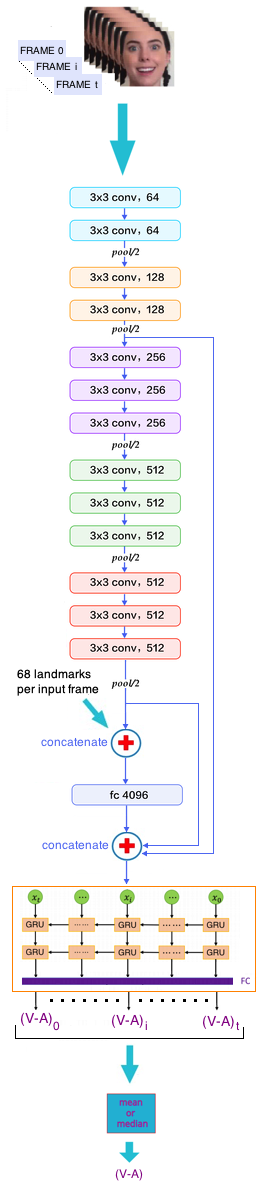}
\caption{The CNN-1RNN-2nd-pool\_last-pool\_fc architecture. It provides a valence-arousal (V-A) estimate per input sequence of consecutive frames. The '68 landmarks' are concatenated with the features of the last 'pool' layer and passed as input to the 'fc' layer.}
\label{cnn_1rnn}
\end{figure}

Fig.\ref{cnn_1rnn} presents one example of CNN-1RNN networks, which we call CNN-1RNN-2nd-pool\_last-pool\_fc. In this network, the features extracted from: i) the second pooling layer (following the fourth convolutional), ii) the last pooling layer (following the 13th convolutional and before being concatenated with the landmarks) and iii) the fully connected layer, are concatenated and passed to the RNN.
As shown in the experimental Section \ref{experiments}, this network 
based on the features extracted from these specific layers provided the best results in these type of networks.

%\subsubsection{CNN-2RNN networks}
%In this architecture, the features extracted from the fully connected layer are passed as input to a RNN network and the features extracted from the  last pooling layer are passed as input to another RNN network. The outputs of those 2 RNNs are concatenated and passed as input either to the output layer (we denote this network as CNN-2RNN) or to a fully connected layer of 100 units followed by the output layer (we denote this network as CNN-2RNN-1FC). 
%Again, for each frame in the input sequence of frames, the CNN-2RNN predicts the valence-arousal values and computes their mean, or median, that are the final estimates.

\subsection{Ensemble Methodology}

In this Subsection we describe an ensemble approach which fuses the developed networks at: i) Model-level and ii) Decision-level. Model-level fusion is based on concatenating the high level features extracted by different networks, whilst Decision-level fusion is based on weighted averaging the predictions provided by different networks. On the one hand side, Model-level fusion takes advantage of the mutual information in the data. On the other hand side, the averaging procedure in Decision-level fusion reduces variance in the ensemble regressor (thus achieving higher robustness), while preserving the relative importance of each individual model. %The main goal of this procedure is to take advantage of the redundancy of a set of independent classifiers to achieve higher robustness by combining their results. 

%\textcolor{red}{Various experiments have been made, considering the different types of networks presented in the previous Subsection. The ones that provided best results are presented next.} -> delete

\subsubsection{Model-level Fusion}

Let us consider the CNN-1RNNs and CNN-3RNNs described in the previous Subsection. We concatenate the outputs of all the RNNs in the above networks and provide them, as input, either: i) to another single RNN layer with 128 GRU units, or ii) to a fully connected layer with 128 units; the output layer follows. We denote the resulting networks as Model-level Fusion + RNN and Model-level Fusion + FC, respectively.  Similarly to the previous Subsections, for each frame in the input sequence of frames, this model-level fusion network predicts the valence-arousal values and then computes their mean, or median, as final estimates. %This network is trained in an end-to-end manner.

\subsubsection{Decision-level Fusion}

%%% isws vale mia eikona me to late fusion

Let us consider again the CNN-1RNNs and CNN-3RNNs described above. The final valence (arousal) estimate $O_{v}^{dec.-level}$ ($O_{a}^{dec.-level}$), is computed as a weighted average of the final valence (arousal) estimates,  $o_{v}^{n} (o_{a}^{n})$, of these networks; each weight is proportional to the corresponding network performance on the validation set:

\begin{equation} \label{eq_1}
%\begin{align} 
{O_{i}}^{dec.-level} =  \frac{1}{\sum\limits_{n}^{}{ t_i^n}} \sum_{n}^{}{ t_i^n} \cdot {o_i^n} 
,  % \nonumber \\  &  \{n,k\} \in\{CNN-1RNN,CNN-3RNN\} %\nonumber \\ &  AffWildNet,DenseNet-RNN \}
%,
%\nonumber \\ &   i \in\{valence,arousal\} \nonumber
%\end{align}
\end{equation}

\noindent where $i \in \{v,a\}$ ($v$ stands for valence, $a$ stands for arousal), 
$t_i^n $ is equal to the Concordance Correlation Coefficient (CCC), $\rho_i$, for valence or arousal, computed on the validation set, with $n$ denoting the CNN-1RNNs or CNN-3RNNs; the CCC  has been the evaluation criterion of the OMG-Emotion Challenge, taking values in $[-1,1]$  and is defined as follows  :

\begin{equation} \label{eq_5}
\rho_i = \frac{2 s_{i,xy}}{s_{i,x}^2 + s_{i,y}^2 + (\bar{x_i} - \bar{y_i})^2},
\end{equation}

\noindent
where $i \in \{v,a\}$, $s_{i,x}$ and $s_{i,y}$ are the variances of the valence/arousal labels and predicted values respectively, $\bar{x_i}$ and $\bar{y_i}$ are the corresponding mean values and $s_{i,xy}$ is the covariance value.

% isws pes pws auto einai kratwntas fixed ta trained independently nets

\section{Network Training Details}\label{training}

In the following, we provide further information regarding the parameters used in the developed architectures (learning rate, dropout probability value, batch size, sequence length), the loss function that was formulated for our problem and the series of post-processing steps that were applied to the obtained estimates of valence and arousal.

\subsection{Implementation Details}

In all developed CNN, CNN plus RNN and CNN plus Multi-RNN architectures, dropout with $0.5$ probability value was applied on the fully connected layers that were on top of the convolutional and pooling layers of CNN networks (VGG-FACE, ResNet-50 and DenseNet-121). Additionally, dropout with $0.8$ probability value was applied after the first GRU layer of the RNNs.

For training our CNN networks, different sequence sizes were used, ranging from 40 to 100, with the size of 80 frames providing the best results. In the CNN plus RNN and CNN plus Multi-RNN cases, we used a batch size of 4 and sequence length of 80 consecutive frames. When training the CNN architectures, the learning rate was chosen to be $10^{-4}$. When training end-to-end the CNN plus RNN architectures, the learning rate was either $10^{-4}$ or $10^{-5}$; when training them, keeping their respective CNN parts fixed, it was $10^{-3}$. All networks were trained using Tensorflow on a Quadro GV100 Volta GPU and the training time was about a day.

\subsection{Objective Function}

Since the evaluation criterion of the OMG-Emotion Challenge was the CCC, our loss function was based on that criterion and was defined as:

\begin{equation} \label{eq:3}
\mathcal{L}_{total} = 1 - \frac{\rho_a + \rho_v}{2},
\end{equation}

\noindent
where $\rho_a$ and $\rho_v$ are the CCC for the arousal and valence.

\subsection{Post-Processing}\label{post-process}

Finally, for all investigated methods, a chain of post-processing steps was applied. These steps included: i) median filtering of the - per frame - predictions within a sequence and ii) smoothing of the - per utterance - predictions (especially to those that consisted of too few frames). Any of these post-processing steps was kept when an improvement was observed on the CCC over the validation set, and applied then, with the same configuration to the test partition.
%% gia to smoothing na to peis etsi i oti to ekanes se per sequences within an utterance?
%% episis isws grapse more post-processing steps opws scaling me mean k std

\section{Experimental Results}\label{experiments}

In all conducted experiments, best results were obtained when the final estimates were the median of the, per frame, valence and arousal estimates within a sequence. %In the following, we focus on the performance obtained when taking the median of the per frame estimates. 
%It should also be stated, that all reported results refer to the test set. 
In all developments, we trained the DNNs with the training set, evaluated them on the respective validation set and selected the best networks according to the validation performance. There were no significant differences between training the DNN multiple times and then averaging the predictions, or using a 10-fold cross validation. %(training on 8 folds, testing on the remaining 2 and in the end averaging the predictions of the networks). 

We examined to include a level of encoding for matching the size of landmarks with the size of the CNN features before fusing them. We first passed the 68 landmarks to a fully connected layer of 512, 1024, or 2048 units and then fused this output with the features extracted from the CNN. However, we did not notice any significant difference in performance, although the developed architectures were more complex and bigger in terms of learnable parameters.

%%% pio stable ta montela apo low level => mikroteri i diafora valid k test performance

%\textcolor{red}{The difference in the trained DNN performance between the validation and the test set was around 3\% for valence and around 5\% for arousal -> auto palia; twra sta CNN-1RNN k CNN-3RNN den poluupirxan.
%This result is probably due to the difference in the test and validation sizes.} As described in Subsection \ref{omg-db}, the validation set consists of 60 videos that contain 617 utterances with a total frame length of around 118,740; the test set consists of 204 videos that contain 2229 utterances of around 431,200 total frames. In other words, the test set is more than 3 times bigger than the validation one. This difference can be further justified by the different distribution of the validation and test set's annotations, as can be seen in Fig.\ref{omg-hist-data}.

\subsection{CNN-RNN Component Analysis}

Table \ref{cnn-results} shows the performance of the developed CNN, standard CNN plus RNN, CNN plus Multi-RNN and ensemble architectures, pre-trained on the Aff-Wild2 database, with and without the post-processing steps described in Subsection \ref{post-process} (for all networks: $p$-value $\leqslant 10^{-20} \ll 0.05$).
The VGG-FACE has achieved the best performance compared to the ResNet-50 and DesNet-121 networks. This is expected as the VGG-FACE network has been pre-trained with a large dataset for face recognition (many human faces have been, therefore, used in its construction), thus better filters are already established in comparison to the ResNet-50 and DesNet-121 that have been pre-trained on objects. Additionally, after further pre-training on Aff-Wild2, a better tuning of these filters is attained in the VGG-FACE case.

Additionally, AffWildNet and DenseNet-RNN networks achieved a better performance than all CNN networks. The former networks are standard CNN plus RNNs in which the RNN is used in order to model the contextual information in the data, taking into account temporal variations and thus a better performance is expected. % isws kapoia exigisi giati AffWildNet outperformed DenseNet-RNN 

One can also note that both CNN-1RNN-2nd-pool\_last-pool\_fc and CNN-3RNN-2nd-pool\_last-pool\_fc exhibit a much improved performance (between 6\% and 10\% on average) when compared to CNN plus RNN architectures. This validates our essence that low-level CNN features together with high-level ones provide useful information for our task. Additionally, CNN-3RNN-2nd-pool\_last-pool\_fc outperformed CNN-1RNN-2nd-pool\_last-pool\_fc showing that it is better to exploit the low- and high-level features' time variations via RNNs, independently, and then concatenate them, rather than concatenate them first and process them through the use of a single RNN. 

%it is more advantageous to pass features through multiple RNN subnets rather than concatenate and pass them through a single RNN. 
%Let us mention here that these models were the ones of the CNN-1RNN and CNN-3RNN types with the best performance and that is why we report on these. + pes oti se both types extracted features from same layers had the best performance

Table \ref{cnn-results} validates that using the ensemble methodology is better than using a single network. This is because different networks produce quite different features; fusing them exploits all these representations that include rich information. It can also be observed that Model-level fusion method has a superior performance compared to that of the Decision-level one, since the features from different networks that are concatenated, contain richer information about the raw data than the final decision. In particular, in Model-level fusion, we concatenate these features and pass them through an RNN and the whole ensemble is trained end-to-end and optimized so that the concatenation of features can provide the best overall result. %In Decision-Level fusion, the ensemble is trained end-to-end and optimized so that the weighted average of each network's score can provide the overall result.}
Moreover, in Model-level fusion, a better performance is achieved when a RNN, instead of a fully connected layer, is used for the fusion.

%\textcolor{red}{Surprisingly enough the late fusion method displayed better results than the early fusion one. Generally, the latter one is expected to have the best performance. We believe that this is so, because the OMG-Emotion dataset is a small one and thus we did not have enough data for jointly training \textcolor{red}{end-to-end} so many deep architectures (with so many parameters). Furthermore, each of the networks that consist the ensemble one have been: i) first pre-trained on the Aff-Wild2 that targets similar task as the OMG dataset and ii) then trained on the OMG dataset and thus our face features %are already derived from a deep pipeline so they
%have already benefited from the generalization properties of deep networks and thus the late fusion works better. -> alliws gia na to apofugeis apla allaxe ta results tou model level k tou late

%The facial landmarks, which are provided as additional input to the network, in this way, contribute to boosting the performance of our model.

%and thus a better performance is acquired.

One can also notice that the post-processing steps helped to achieve a better performance, mainly in valence estimation. The median filter size that we used was 81 for valence (similar to the sequence length), whereas only 3 for the arousal.  % kanonika sto 20 itan gia valence -to eida molis-, gia arousal den to thumamai 
The arousal window size was small, but, when it was increased, the performance decreased. %\textcolor{red}{This in essence means that for the frames within an utterance, the emotional state itself did not change, but the intensity did change.}
Our final observation is that the performance of the networks in arousal was worse than their performance in valence. This is expected because we only used the visual modality for training our networks; for arousal the audio cues appear to include more discriminating capabilities than facial features in terms of correlation coefficient; this conclusion confirms previous findings \cite{nicolaou2011continuous}.

\begin{table}[ht]
\caption{CCC based evaluation, on the OMG test set, of valence \& arousal predictions provided by our developed CNN, CNN plus RNN, CNN plus Multi-RNN and ensemble architectures. All networks are pre-trained on Aff-Wild2 with (without) post-processing. A higher CCC value indicates a better performance.}
\label{cnn-results}
\centering
\scalebox{0.87}{
\begin{tabular}{ |c||c|c|c| }
 \hline
 \multicolumn{1}{|c||}{CCC} & \multicolumn{2}{c|}{ \begin{tabular}{@{}c@{}} With (Without) \\ Post-Processing \end{tabular}} & \multicolumn{1}{c|}{ \begin{tabular}{@{}c@{}} Mean \end{tabular}}  \\
  \hline
  & Valence & Arousal &  \\
 \hline
\textit{VGG-Face} & \textit{0.378} \textit{(0.361)} & \textit{0.203} \textit{(0.193)} & \textit{0.291} \textit{(0.277)}   \\
\hline
 DenseNet-121 & 0.365 (0.350) & 0.191 (0.184) & 0.278 (0.267)  \\
 \hline
 ResNet-50 & 0.359 (0.344) & 0.195 (0.189) & 0.277 (0.267) \\
 \hline
  \textit{AffWildNet} & \textit{0.409} \textit{(0.390)} & \textit{0.224} \textit{(0.219)} & \textit{0.317} \textit{(0.305)}  \\
\hline
 DenseNet-RNN &0.394 (0.378) &0.211 (0.209) & 0.303 (0.294)   \\
 \hline
 \begin{tabular}{@{}c@{}} \textit{CNN-1RNN-} \\ \textit{2nd-pool\_last-pool\_fc} \end{tabular} & \textit{0.449} \textit{(0.441)}  &\textit{0.303} \textit{(0.297)}   & \textit{0.376} \textit{(0.369)}      \\
 \hline
  \begin{tabular}{@{}c@{}} \textbf{CNN-3RNN-} \\ \textbf{2nd-pool\_last-pool\_fc} \end{tabular} &\textbf{0.472} \textbf{(0.463)}  & \textbf{0.329} \textbf{(0.322)}  & \textbf{0.401} \textbf{(0.393)}     \\
 \hline
 Decision-Level Fusion & 0.501 (0.482) & 0.332 (0.321) & 0.417 (0.402)  \\
\hline
 \textit{Model-Level Fusion + FC} &  \textit{0.518}  \textit{(0.500)} &  \textit{0.348}  \textit{(0.328)} &  \textit{0.433}  \textit{(0.414)} \\
 \hline
 \textbf{\textit{Model-Level Fusion + RNN}} &  \textbf{\textit{0.535}}  \textbf{\textit{(0.512)}} &  \textbf{\textit{0.365}}  \textbf{\textit{(0.340)}} &  \textbf{\textit{0.450}}  \textbf{\textit{(0.426)}} \\
 \hline
\end{tabular}
}
\end{table}

%Additionally, comparing Tables \ref{cnn-results} and \ref{cnn-rnn-results}, it can be verified, as expected, that the CNN plus RNN architectures, modelled the temporal evolution in the best way, providing higher performance in valence and arousal estimation than the CNN ones. Again, the post-processing boosted the performance of the networks.

In the following, we compare the performance of the best performing networks of Table \ref{cnn-results} with post-processing to that of networks trained from scratch, or being pre-trained with the Aff-Wild or the Aff-Wild2 database. Table \ref{final} presents the results of this comparison. %It can be seen that all  models pre-trained on the Aff-Wild2 database outperform all other models in both valence and arousal estimation. 
The Aff-Wild2 database, due to its big size and emotion diversity, boosted the performance of all networks pre-trained with it, in comparison to the performance of the networks trained directly with the OMG-Emotion set. This was also the case when we pre-trained the networks with the Aff-Wild database. Overall, networks pre-trained with the Aff-Wild2 achieved a better performance in comparison to networks pre-trained with the Aff-Wild database.

\begin{table}[ht]
\caption{CCC based evaluation, on the OMG test set, of valence \& arousal predictions provided by various networks when: they are trained from scratch or are pre-trained with the Aff-Wild and Aff-Wild2  databases. A higher CCC value indicates a better performance.}
\label{final}
\centering
\scalebox{0.7}{
\begin{tabular}{ |c||c|c|c|c|c|c| }
 \hline
\multicolumn{1}{|c||}{Methods} & \multicolumn{2}{c|}{\begin{tabular}{@{}c@{}}Trained \\  from Scratch \end{tabular}} & \multicolumn{2}{c|}{\begin{tabular}{@{}c@{}}  Pre-trained \\ on Aff-Wild \end{tabular}}  & \multicolumn{2}{c|}{\begin{tabular}{@{}c@{}}  Pre-trained \\ on  Aff-Wild2 \end{tabular}}  \\
 \hline
  & Valence & Arousal & Valence & Arousal & Valence & Arousal \\
%\hline
%VGG-FACE  & 0.312  & 0.147  & 0.349  & 0.177 & \textbf{0.378}& \textbf{0.203}     \\
\hline
\begin{tabular}{@{}c@{}} CNN-1RNN- \\ 2nd-pool\_last-pool\_fc \end{tabular} & 0.371  & 0.210 & 0.419  & 0.278    &\textbf{0.449}  & \textbf{0.303}     \\
\hline
\begin{tabular}{@{}c@{}}CNN-3RNN- \\ 2nd-pool\_last-pool\_fc\end{tabular}  & 0.385   & 0.192   & 0.448  & 0.302  &\textbf{0.472}  &\textbf{0.329}   \\  
 \hline
Model-level Fusion + RNN  & 0.431 & 0.265  & 0.511  & 0.342  & \textbf{0.535} & \textbf{0.365}   \\  
\hline
\end{tabular}
}
\end{table}

Between CNN-1RNN and CNN-3RNN types of architectures, a better performance was acquired when using the latter one. Next, we present an ablation study on extracting different CNN low-, mid- and high-level features in CNN-3RNN networks. Table \ref{effect} compares their performance (in all cases: $p$-value $\leqslant 10^{-25} \ll 0.05 $). The first four rows of Table \ref{effect} show the performance of networks where a combination of low-, mid- and high-level features are extracted, whereas the next rows show the performance of networks where only low-, or only mid-, or only high-level features are extracted.
Let us note that worst performances among all these types of networks were obtained when features were extracted from mid- CNN levels (convolutional layers 6-9).
Generally, best performances were obtained when features were extracted from high- and from low-levels. The optimal combination (that provided the best performance) was through the use of CNN-3RNN-2nd-pool\_last-pool\_fc. One more observation is that low-level features (convolutional layers 3-5), especially when combined with high-level, significantly affected the performance in predicting both valence and arousal.

\begin{table}[ht]
\caption{Effect on CCC (on the OMG test set) of using features from different layers in the CNN-3RNN case. All networks are post-processed \& pre-trained on Aff-Wild2. A higher CCC value indicates a better performance.}
\label{effect}
\centering
\begin{tabular}{ |c||c|c|c| }
 \hline
 \multicolumn{1}{|c||}{CNN-3RNN} & \multicolumn{2}{c|}{ \begin{tabular}{@{}c@{}} CCC \end{tabular}} & \multicolumn{1}{c|}{ \begin{tabular}{@{}c@{}} Mean \end{tabular}}  \\
  \hline
  & Valence & Arousal &  \\
%\hline
%VGG-FACE  & 0.312  & 0.147  & 0.349  & 0.177 & \textbf{0.378}& \textbf{0.203}     \\
\hline
8th conv + last pool + fc  & 0.416   & 0.261   & 0.339    \\  
\hline
5th conv + last pool + fc  & 0.455   & 0.322   & 0.389    \\  
\hline
2nd pool + last pool + fc & \textbf{0.472}   & \textbf{0.329}   & \textbf{0.401}    \\  
 \hline
3rd conv + 7th conv + fc  & 0.402   & 0.267   & 0.335    \\ 
\hline
last conv + last pool + fc & 0.440  & 0.248 & 0.344   \\
 \hline
6th conv + 7th conv + 8th conv  & 0.328   & 0.162   & 0.245    \\ 
 \hline
7th conv + 8th conv + 9th conv  & 0.334   & 0.172   & 0.253    \\ 
 \hline
3rd conv + 4th conv + 5th conv  & 0.345   & 0.185   & 0.265    \\ 
 \hline
\end{tabular}
\end{table}

%In an effort to clarify what the obtained performance values for valence and arousal mean and for additional insight concerning the problem, Figs. \ref{hist_lab_pred_v} and \ref{hist_lab_pred_a} display the predicted values and labels of valence and arousal, respectively, of the whole test set of the Challenge.

Next, we present an ablation study on the use of landmarks as additional input to various networks. Table \ref{landmarks} compares the performance of the CNN-1RNN-2nd-pool\_last-pool\_fc,  CNN-3RNN-2nd-pool\_last-pool\_fc and Model-level Fusion + RNN networks when the landmarks are and are not used as additional input. In all cases, using landmarks increases their performance by 1.2\% - 1.9\%.

\begin{table}[ht]
\caption{Effect on CCC (on the OMG test set) of (not) using landmarks as additional input to various networks. All networks are post-processed \& pre-trained on Aff-Wild2. A higher CCC value indicates a better performance. V,A stand for Valence and Arousal}
\label{landmarks}
\centering
\scalebox{0.82}{
\begin{tabular}{ |c||c|c|c|c|c|c| }
 \hline
\multicolumn{1}{|c||}{CCC} & \multicolumn{3}{c|}{\begin{tabular}{@{}c@{}} Without Landmarks \end{tabular}} & \multicolumn{3}{c|}{\begin{tabular}{@{}c@{}}  With Landmarks \end{tabular}}  \\
 \hline
  & V & A & Mean & V & A & Mean  \\
\hline
\begin{tabular}{@{}c@{}} CNN-1RNN- \\ 2nd-pool\_last-pool\_fc \end{tabular} & 0.429 & 0.291 & 0.360  &\textbf{0.449}  &\textbf{0.303}  &\textbf{0.376}    \\
\hline
\begin{tabular}{@{}c@{}}CNN-3RNN- \\ 2nd-pool\_last-pool\_fc \end{tabular} & 0.454  & 0.310  & 0.382  &\textbf{0.472}  &\textbf{0.329} & \textbf{0.401}   \\  
 \hline
Model-level Fusion + RNN  & 0.524  & 0.352  & 0.438 &\textbf{0.535}  &\textbf{0.365} & \textbf{0.450}   \\  
\hline
\end{tabular}
}
\end{table}

Finally, to give more insight on the performance of the best CNN-3RNN (CNN-3RNN-2nd-pool\_last-pool\_fc), we analyzed its performance at different parts of the 2D Valence-Arousal Space. Table \ref{mse} presents the obtained valence and arousal performance in terms of Mean Squared Error (MSE) across 4 different regions of this Space. It can be seen that better results have been obtained in the region with high arousal and positive valence; however the obtained MSE are not far away from the MSE across the whole 2D Valence-Arousal Space.

\begin{table}[ht]
\caption{Valence and Arousal MSE in areas of the 2D VA Space for the best CNN-3RNN. A lower MSE indicates a better performance. V,A stand for Valence and Arousal}
\label{mse}
\centering
\scalebox{0.64}{
\begin{tabular}{ |c||c|c|c|c|c| }
 \hline
\multicolumn{1}{|c||}{2D VA-Space} & \multicolumn{1}{c|}{\begin{tabular}{@{}c@{}} V $\in$ [0,1] \\ A $\in$ [0,0.5)  \end{tabular}} & \multicolumn{1}{c|}{\begin{tabular}{@{}c@{}}  V $\in$ [0,1] \\ A $\in$ [0.5,1]  \end{tabular}} & \multicolumn{1}{c|}{\begin{tabular}{@{}c@{}}  V $\in$ [-1,0) \\ A $\in$ [0,0.5)  \end{tabular}} & \multicolumn{1}{c|}{\begin{tabular}{@{}c@{}}  V $\in$ [-1,0) \\ A $\in$ [0.5,1] \end{tabular}} & \multicolumn{1}{c|}{\begin{tabular}{@{}c@{}}  V $\in$ [-1,1] \\ A $\in$ [0,1] \end{tabular}} \\
\hline
\hline
\begin{tabular}{@{}c@{}} CNN-3RNN- \\ 2nd-pool\_ \\last-pool\_fc \end{tabular} & {\begin{tabular}{@{}c@{}} \\  MSE-V = 0.101 \\ MSE-A = 0.031  \end{tabular}}  & {\begin{tabular}{@{}c@{}} \\  MSE-V = 0.055 \\ MSE-A = 0.021  \end{tabular}} & {\begin{tabular}{@{}c@{}} \\ MSE-V = 0.154 \\ MSE-A = 0.061 \end{tabular}} & {\begin{tabular}{@{}c@{}} \\ MSE-V = 0.110 \\ MSE-A = 0.040  \end{tabular}} & {\begin{tabular}{@{}c@{}} \\  MSE-V = 0.110 \\ MSE-A = 0.041  \end{tabular}}  \\  
\hline
\end{tabular}
}
\end{table}

%\begin{figure}[h]
%\centering
%\adjincludegraphics[height=7cm,width=9.5cm]{figures/test_lab_pred_valence.png}
%\caption{Valence predictions versus labels of the test set of the OMG-Emotion Challenge}
%\label{hist_lab_pred_v}
%\end{figure}

%%%% isws ftiaxe ligo tis eikones autes

%\begin{figure}[h]
%\centering
%\adjincludegraphics[height=7cm,width=9.5cm]{figures/test_lab_pred_arousal.png}
%\caption{Arousal predictions versus labels of the test set of the OMG-Emotion Challenge}
%\label{hist_lab_pred_a}
%\end{figure}

\subsection{Submissions to the OMG-Emotion Challenge}

For the OMG-Emotion Challenge each team was allowed to have up to 3 submissions. We have submitted the CNN2-to-1RNN and CNN-3RNN-last-conv\_last-pool\_fc  pre-trained on Aff-Wild models' predictions without post-processing (submission I) and with post-processing: either with median filtering (submission II) or with median filtering and smoothing (submission III; our best one). More details regarding our submissions can be found in \cite{kollias2018old}.

\subsection{Comparison with State-of-the-Art}

Here we compare the performance of our best networks to the performances of state-of-the-art methods submitted to the OMG-Emotion Challenge. The authors of \cite{peng2018deep} developed the VNet and ANet models. VNet is a SphereFace \cite{liu2017sphereface} network, followed by a BLSTM, followed by a temporal pooling and the output layer. ANet is a VGG16 network with average pooling and accepts as input STFT maps extracted from the audio. In their fusion, the features extracted from VNet's temporal pooling and ANet's average pooling layers, are concatenated and passed to the output layer.

The authors of \cite{triantafyllopoulos2018audeering} developed two models. In the first model, denoted as openSMILE + LSTMs, features extracted from audio using openSMILE \cite{eyben2010opensmile} were passed through six 2-layer LSTMs, each predicting valence, arousal or both; the final prediction was their average. In the second model, denoted as VGG-FACE-BLSTM, the visual modality was used; frames from the utterances were passed through a fixed and pre-trained VGG-FACE followed by a 2-layer BLSTM that gave the final valence prediction. %In all submissions, the arousal was computed from the audio. In the first submission, denoted as Sub I, valence was computed from the video, whereas in the second one, denoted as Sub II, valence was computed from both audio and video (being their average), whereas in the third submission, denoted as Sub III, valence was computed from the audio.

The authors of \cite{zheng2018multimodal} developed both single and ensemble networks, consisting of three models. In the first model, denoted as Single Multi-Modal, acoustic features were extracted using openSmile; visual features were extracted from a fixed and pre-trained VGG16 followed by 1-layer LSTM with attention mechanism; visual and acoustic features were passed into an SVM that performed the final predictions. The second model was similar to the first and extracted similar visual and acoustic features, but it also extracted acoustic features from SoundNet. All these features were passed to an SVM that performed the predictions. The late fusion of the two afore-mentioned models, is denoted as Ensemble I; the final predictions were a weighted sum of the models' predictions. The third model was an end-to-end trained VGG16 followed by 1-layer LSTM with attention mechanism that takes as input only visual data. The late fusion of the three developed models, is denoted as Ensemble II; again the final predictions were a weighted sum of the models' predictions.

%late fusion of two networks is performed: the one network is the Single Multi-Modal described before; 
%the second network is similar to the first and extracts similar visual and acoustic features, but it also extracts acoustic features from SoundNet. All these features are passed into an SVM that performs the predictions.
%The final predictions are a weighted sum of these two models' predictions.
%The third model, denoted as Ensemble II, consists of three models. The first two models are the Single Multi-Modal and Ensemble I. The third model is an end-to-end trained VGG16 followed by 1-layer LSTM with attention mechanism that takes as input only visual data. The final predictions are a weighted sum of these three models' predictions.

Table \ref{sota} shows that our Model-level Fusion + RNN method outperforms all other methods -even those that have been trained using the audio modality as well- on both the valence and arousal estimation. Table \ref{sota} also shows that the CNN-3RNN-2nd-pool\_last-pool\_fc outperformed all state-of-the-art networks, regardless whether they additionally used the audio modality, except for: i) the Single Multi-Modal method that outperformed it on average by 0.015 (however this network used the audio modality as well; since the audio and speech contribute more to arousal estimation, this small difference is justified) and ii) Ensembles I and II, which are a fusion of many different networks that used the visual and audio modalities and thus again the difference in performance was expected. 
% For arousal, the  audio  cues  appear  to  perform  better  than  the  facial features  in  terms  of  correlation  coefficient;  this  conclusion confirms previous findings \cite{nicolaou2011continuous}.

\begin{table}[ht]
\caption{CCC based evaluation, on the OMG test set, of VA predictions provided by our best performing networks vs the state-of-the-art. %that used different modalities 
V,A stand for valence and arousal. A higher CCC value indicates a better performance. } %the results are taken from \url{https://www2.informatik.uni-hamburg.de/wtm/omgchallenges/omg_emotion2018_results2018.html}}
\label{sota}
\centering
\scalebox{0.89}{
\begin{tabular}{ |c||c|c|c| }
 \hline
\multicolumn{1}{|c||}{Methods} &
%\multicolumn{1}{c||}{Team} & 
\multicolumn{1}{c|}{\begin{tabular}{@{}c@{}} Modality \end{tabular}} & \multicolumn{2}{c|}{\begin{tabular}{@{}c@{}}  CCC \end{tabular}}   \\
 \hline
 & & Valence & Arousal \\ 
  \hline
  VNet \cite{peng2018deep}  & V,A: visual & 0.438 & 0.244     \\
\hline
  ANet + VNet  \cite{peng2018deep}  & V,A: audio + visual & 0.442 & 0.236     \\
\hline
\begin{tabular}{@{}c@{}}  openSMILE + LSTMs, \\ VGG-FACE-BLSTM \end{tabular} \cite{triantafyllopoulos2018audeering} & \begin{tabular}{@{}c@{}} A: audio, \\ V: visual \end{tabular} &	0.258	& 0.277    \\
\hline
\begin{tabular}{@{}c@{}} openSMILE + LSTMs, \\ VGG-FACE-BLSTM + \\ openSMILE + LSTMs \end{tabular}  \cite{triantafyllopoulos2018audeering} & \begin{tabular}{@{}c@{}} A: audio, \\ V: audio + visual \end{tabular} &	0.369 & 0.286  \\
\hline
 openSMILE + LSTMs \cite{triantafyllopoulos2018audeering} & V,A: audio & 0.361 & 	0.293    \\
\hline
\begin{tabular}{@{}c@{}} Single Multi-Modal \cite{zheng2018multimodal} \end{tabular}  & V,A: audio + visual   &	0.484 & 0.345  \\  
 \hline
\begin{tabular}{@{}c@{}} Ensemble  I \cite{zheng2018multimodal} \end{tabular}   & V,A: audio + visual   &	0.496 & 0.356   \\  
 \hline
\begin{tabular}{@{}c@{}} Ensemble  II \cite{zheng2018multimodal} \end{tabular}   & V,A: audio + visual   & 0.499  & 0.361   \\
 \hline
%Trimodal \cite{deng2018multimodal}  & V,A: audio + visual + text & 0.359 & 0.277  \\  
\hhline{=:=:=:=}
\begin{tabular}{@{}c@{}}CNN-3RNN- \\ 2nd-pool\_last-pool\_fc\end{tabular}  & V,A: visual & 0.472 & 0.329  \\  
\hline
Model-level Fusion + RNN   & V,A: visual & \textbf{0.535} & \textbf{0.365}  \\  
\hline
\end{tabular}
}
\end{table}

%\subsection{Discussion}

%The results provided in this Section, confirmed five intuitions that we had. 

%\textcolor{red}{First, there is a performance drop from the validation to the test set because of their great difference in size.} 

%Secondly, the CNN plus RNN architectures exhibit better performance than the CNN architectures and the ensemble ones have better performance than all the single ones. 

%Thirdly, the Aff-Wild2 database, due to its big size and emotion diversity, boosted the performance of all the networks when we pre-trained them on it, in comparison to the case that we trained them directly on the OMG-Emotion training set. This was also the case when we pre-trained the networks on the Aff-Wild database. Overall, networks, when pre-trained on the Aff-Wild2, showed a better performance in comparison to the case where they were pre-trained on the Aff-Wild database.

%Additionally, the post-processing steps helped into achieving a bigger performance. 

%Lastly, the performance of the networks in arousal was worse than their performance in valence, because for arousal the audio cues appear to perform better than the facial features in terms of correlation coefficient; this conclusion confirms previous findings \cite{nicolaou2011continuous}.

\section{Conclusions}\label{conclusion}

This paper presented the development of novel architectures for predicting valence-arousal, by utilizing the OMG-Emotion dataset. The proposed approach was based on visual information and achieved very good performance when tested on the OMG-Emotion test set.
In the developed networks, features extracted from low-, mid- and high- CNN layers were either concatenated and fed to a single RNN, or processed by RNN subnets and then concatenated. %The output of those, constituted intermediate level predictions that were averaged across input sequences of frames in order to give the final valence-arousal estimates. %Ensemble methodology, given the developed networks, was also used. 
Moreover an ensemble approach was proposed; the Model-level fusion through a RNN produced the best results.
All developed networks were first pre-trained on the rich and large Aff-Wild or Aff-Wild2 databases.

%In the paper, we also described the architectures that were used in our submission to the OMG-Emotion Challenge, through which we ranked second in valence estimation among all contestants that used only visual information, or we ranked third among all contestants that used all available multi-modal information.

\ifCLASSOPTIONcompsoc
  \section*{Acknowledgments}
\else
  \section*{Acknowledgment}
\fi

The work of Dimitris Kollias was funded by a Teaching Fellowship of Imperial College London. Additionally, we would like to thank the reviewers for their valuable comments that helped us to improve this paper.

\bibliographystyle{spmpsci}      
\bibliography{sample}


\begin{IEEEbiography}[{\includegraphics[width=1in,height=1.25in,clip,keepaspectratio]{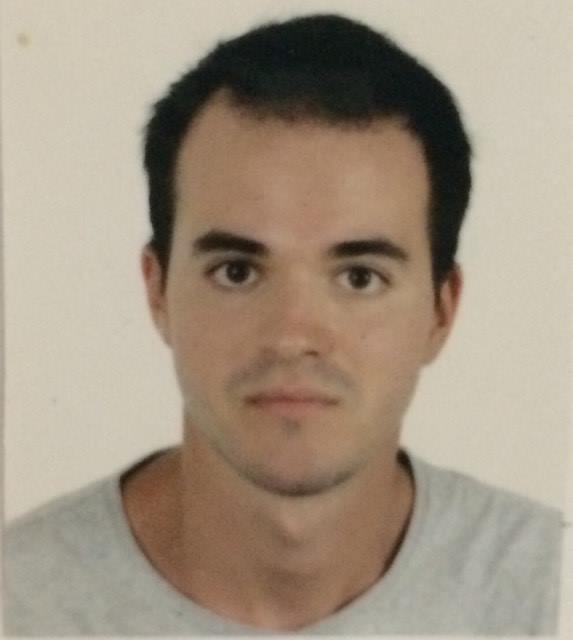}}]{Dimitrios Kollias}, Fellow of the Higher Education Academy (HEA), holder of a PostGraduate Certificate (PG Cert) and student member of the IEEE, received the Diploma/M.Sc. in Electrical and Computer Engineering from the ECE School of the National Technical University of Athens, Greece, in 2015 and the 
M.Sc. in Advanced Computing  from the Department of Computing of Imperial College
London, U.K., in 2016. He is currently working
towards the Ph.D. degree in the intelligent Behaviour Understanding Group (iBUG), having received the prestigious Teaching Fellowship of Imperial College London. During the course of his
Ph.D., he has published his research in the top venues for machine learning, perception and computer vision such as the International Journal of Computer Vision, the CVPR, ECCV and BMVC Conferences, the IEEE IJCNN Conference and the IEEE SSCI. His research interests span the areas of deep
learning and deep neural networks, computer vision, affective computing and medical imaging. 
\end{IEEEbiography}

\begin{IEEEbiography}[{\includegraphics[width=1in,height=1.25in,clip,keepaspectratio]{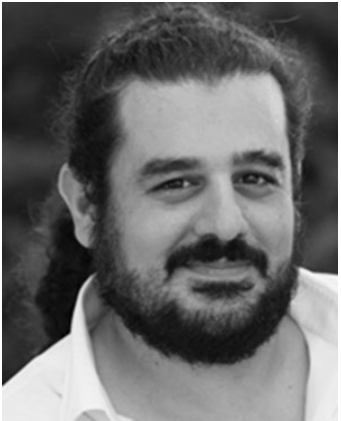}}]{Stefanos Zafeiriou} is currently a Reader in Machine Learning and Computer Vision with the Department of Computing, Imperial College London, and a Distinguishing Research Fellow with the University of Oulu. He received the Prestigious Junior Research Fellowships from Imperial College London in 2011 to start his own independent research group. He received the President’s Medal for Excellence in Research Supervision for 2016. He received various awards during his doctoral and postdoctoral studies. He has been a Guest Editor of more than 6 journal special issues and co-organized more than 15 workshops/special sessions on specialized computer vision topics in top venues, such as CVPR/FG/ICCV/ECCV. He has coauthored more than 70 journal papers mainly on novel statistical machine learning methodologies applied to computer vision problems, such as 2-D/3-D face analysis, deformable object fitting and
tracking, shape from shading, and human behavior analysis, published in the
most prestigious journals in his field of research, such as IEEE TPAMI,
IJCV, IEEE TIP, IEEE TNNLS and many papers in top conferences, such
as CVPR, ICCV, ECCV, ICML. His students are frequent recipients of very
prestigious and highly competitive fellowships, such as the Google, Intel and Qualcomm ones. He has more than
7200 citations to his work, h-index 44. He was the General Chair of BMVC 2017.
\end{IEEEbiography}

\end{document}